\documentclass[10pt,twocolumn,letterpaper]{article}

\usepackage{cvpr}              

\usepackage{graphicx}
\usepackage{amsmath}
\usepackage{amssymb}
\usepackage{booktabs}

\usepackage[pagebackref,breaklinks,colorlinks]{hyperref}

\usepackage[capitalize]{cleveref}
\crefname{section}{Sec.}{Secs.}
\Crefname{section}{Section}{Sections}
\Crefname{table}{Table}{Tables}
\crefname{table}{Tab.}{Tabs.}

\def\cvprPaperID{*****} 
\def\confName{CVPR}
\def\confYear{2025}

\begin{document}

\title{Investigating Vision-Language Model for Point Cloud-based Vehicle Classification}
\author{Yiqiao Li, Jie Wei, Camille Kamga\\
City College of New York\\
160 Convent Ave, New York, NY 10031\\
{\tt\small \{yli4, jwei, ckamga\}@ccny.cuny.edu}
}
\maketitle

\begin{abstract}
Heavy-duty trucks pose significant safety challenges due to their large size and limited maneuverability compared to passenger vehicles. A deeper understanding of truck characteristics is essential for enhancing the safety perspective of cooperative autonomous driving. Traditional LiDAR-based truck classification methods rely on extensive manual annotations, which makes them labor-intensive and costly. The rapid advancement of large language models (LLMs) trained on massive datasets presents an opportunity to leverage their few-shot learning capabilities for truck classification. However, existing vision-language models (VLMs) are primarily trained on image datasets, which makes it challenging to directly process point cloud data. This study introduces a novel framework that integrates roadside LiDAR point cloud data with VLMs to facilitate efficient and accurate truck classification, which supports cooperative and safe driving environments. This study introduces three key innovations: (1) leveraging real-world LiDAR datasets for model development, (2) designing a preprocessing pipeline to adapt point cloud data for VLM input, including point cloud registration for dense 3D rendering and mathematical morphological techniques to enhance feature representation, and (3) utilizing in-context learning with few-shot prompting to enable vehicle classification with minimally labeled training data. Experimental results demonstrate encouraging performance of this method and present its potential to reduce annotation efforts while improving classification accuracy.
\end{abstract}

\vspace*{-2mm}
\section{Introduction}
\label{sec:intro}
Heavy trucks are vital components of roadway traffic but present significant safety challenge due to their size, weight, and limited maneuverability. Such vehicles require longer stopping distances compared to passenger cars and have wider blind spots\cite{jansen2022caught}, which increases the potential for severe collision. A comprehensive understanding of truck characteristics, such as body configuration and movement patterns, is essential for improving traffic safety, designing better infrastructure, and subsequently enhancing cooperative driving systems.

Traditional truck classification methods using LiDAR sensors rely on hand-crafted feature extraction and extensive manual annotations to establish robust datasets\cite{li2021truck}\cite{li2023lidar}. These approaches are time-consuming, expensive, and often lack generalizability across different road environments. Recent advancements in Multi-modal Large Langrage Model, particularly vision-language models (VLMs), have shown remarkable performance in various image-based tasks by leveraging large-scale pretraining and few-shot learning capabilities \cite{huang2023chatgpt}. However, most VLMs are trained on 2D image datasets, which poses a significant challenge when applying them to the point cloud data.
This study aims to address the existing gap by leveraging the representational power of VLMs for LiDAR-based heavy-duty truck classification. The key contributions of this work are as follows. First, We utilize roadside LiDAR sensor data to capture detailed point cloud representations of heavy-duty trucks, ensuring the approach's practical applicability in real-world scenarios.
Second, we propose a systematic method to adapt point cloud data for VLM input. This includes point cloud registration to generate dense 3D renderings and point cloud smoothing techniques to enhance feature representation, improving the model's ability to process and classify the data. Third, we introduce a few-shot prompting approach that allows VLMs to classify vehicles, particularly heavy-duty trucks, without costly parameter updates. This approach significantly reduces the need for extensive manual annotations, which makes the classification process more efficient and scalable.
\section{Preliminary}
\subsection{Vision Language Model}
Vision-language models (VLMs) are a class of multimodal generative models designed to process and understand both visual and textual data. These models take image and text inputs and generate text-based outputs, which enables a wide range of applications. Large VLMs demonstrate strong zero-shot performance and generalize effectively across diverse image types—including documents, web pages, and photographs—and support tasks such as image-based chat, instruction-driven recognition, visual question answering, document understanding, and image captioning \cite{lin2024vila}. Some advanced VLMs, e.g., DeepSeek-VL, also incorporate spatial reasoning, allowing them to detect, segment, and localize objects within an image \cite{lu2024deepseek}. When prompted, they can generate bounding boxes or segmentation masks, identify specific subjects, and answer questions about spatial relationships. The capabilities of VLMs vary significantly based on their training data, image encoding strategies, and architectural design, which leads to diverse strengths across different applications.

\subsection{In-Context Learning}
\paragraph{In-Context Learning} 
In-context learning (ICL) \cite{jiang2024many} is a prompt engineering technique in which task demonstrations are embedded within the input prompt in natural language, which allows the model to infer the desired task without explicit parameter updates. This method enables the adoption of pre-trained VLMs
for novel tasks without costly fine-tuning.

\paragraph{In-Context Learning with few-shot demonstrations} 
ICL with few-shot demonstrations, also known as few-shot prompting \cite{brown2020language}\cite{dong2022survey}, is a prominent approach for multi-class classification using VLMs. In the context of VLM-based multi-class classification, this problem can be framed as follows: Given a query input tokenized image $x$ and a set of candidate classes $Y = {y_1, ..., y_n}$, a pretrained VLM $V$ predicts the answer with the highest prediction score. This prediction is based on a demonstration set $E$ which consists of an optional task instruction $I$ and $k$ demonstration examples. Therefore, $E$ can be represented in two possible ways: $E = \{I, u(x _1,y_1),...,u(x_k,y_k)\}$ or $E = \{u'(x_1,y_1,I),...,u'(x_k,y_k,I)\}$, where $u'(x_1,y_1,I)$ represents an image example tailored to the task. The likelihood of each candidate answer $y_i$ is determined by a scoring function $f$, which evaluates the entire input sequence. This setup allows the model to choose the optimal predictions by considering the input image, a few demonstration examples, and the task instruction. 
\begin{equation}
    P(y_i \mid x) \stackrel{\Delta}{=} f_V(y_i, F, x)
    \label{eq:incontext}
\end{equation}

The final prediction can be written as an argument of the maximum of the conditional probability as fellows:

\begin{equation}
    \hat{y} = \text{arg max}\,_{y_i\in Y}\ P(y_i|x)  
    \label{eq:incontext_prob}
\end{equation}

\section{Vision Language Model (VLM) for Point Cloud-based Truck Classification}
\subsection{Point Cloud Data Processing}
This study adopts the infrastructure-based LiDAR data processing pipeline from \cite{li2021truck}. Vehicle point clouds were first segmented using a background subtraction method that divided the LiDAR sensor’s conical surface into annular sector-shaped cells, isolating foreground vehicles based on spatial occupancy. DBSCAN clustering then grouped points into distinct vehicle objects\cite{ester1996density}. For cross-frame tracking, the SORT algorithm was applied, representing each vehicle by the centroid of its minimum oriented 2D bounding box \cite{bewley2016simple}. Inter-frame displacement was estimated using a Kalman filter, with vehicle assignments optimized via the Hungarian algorithm.

Each individual LiDAR frame is too sparse/scattered to accurately capture the configuration of vehicle objects, especially when compared to RGB image-based methods. To create a better 2D rendering of the point cloud for input into the vision-language model (VLM), which is primarily trained on RGB images, this study adopted a vehicle point registration framework to enhance the resolution of the point cloud images \cite{li2023lidar}.

\begin{figure}[h]
    \centering
    \begin{subfigure}[b]{0.45\textwidth}
        \centering
        \includegraphics[width=\textwidth]{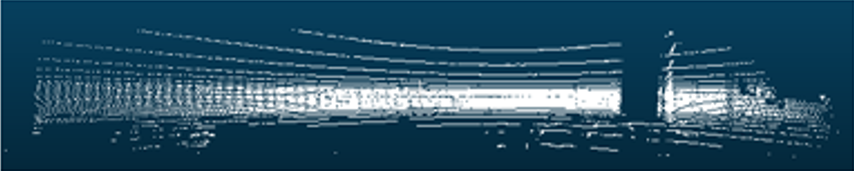}
        \caption{Single Frame}
        \label{fig:single_frame}
    \end{subfigure}
    \hfill
    \begin{subfigure}[b]{0.45\textwidth}
        \centering
        \includegraphics[width=\textwidth]{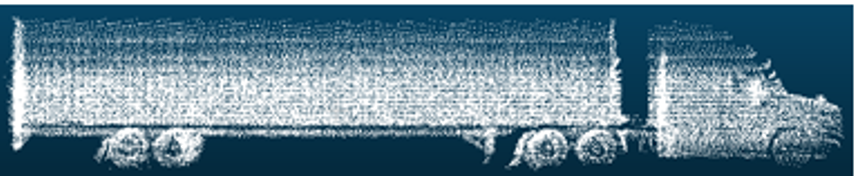}
        \caption{Reconstructed Frame}
        \label{fig:reconstruct}
    \end{subfigure}
    \caption{Comparison of single frame and reconstructed frame.}
    \label{fig:frame_comparison}
\end{figure}

A probabilistic-based pairwise point cloud registration approach was applied to align vehicle objects between consecutive frames\cite{gao2019filterreg}. First, vehicle objects from adjacent frames were aligned by minimizing point-to-point distances. This registration was further refined through a point-to-plane strategy, which enhances the precision of vehicle point cloud alignment, particularly when a well-defined surface is established as the vehicle approaches the LiDAR sensors, where the point-to-point method may become less effective. Finally, single-frame vehicle point clouds were reconstructed using the transformation matrices derived from consecutive frames, which improves their resolution and produces a more detailed 3D representation. A visual comparison between a single frame of a tractor-trailer truck and the reconstructed results is presented in Figure \ref{fig:frame_comparison}. The reconstructed truck provides a clearer definition of the vehicle's edges in the point cloud compared to the single frame results, which offers a more precise representation that enhances the VLM model's ability to interpret and perform classification tasks effectively.

\subsection{Point Cloud Image Processing}
To optimize point cloud images used as input to VLMs, this study applied {\it statistical outlier removal}\cite{zhou2018open3d} and mathematical morphological operators - {\it opening} ( {\it Erosion} followed by {\it Dilation})\cite{najman2013mathematical}. These techniques help refine and smooth the contours of foreground objects, which effectively eliminate small noise both from the point cloud and its 2D projections. By mitigating this noise while preserving the overall structure of the vehicle, the process results in a cleaner, more continuous 2D representation of the point cloud. This improved representation enhances the suitability of the point cloud for the classification task in VLM-based applications.

\subsection{Few-shot Prompting}
\begin{figure}[h]
    \centering
    \includegraphics[width=0.5\textwidth]{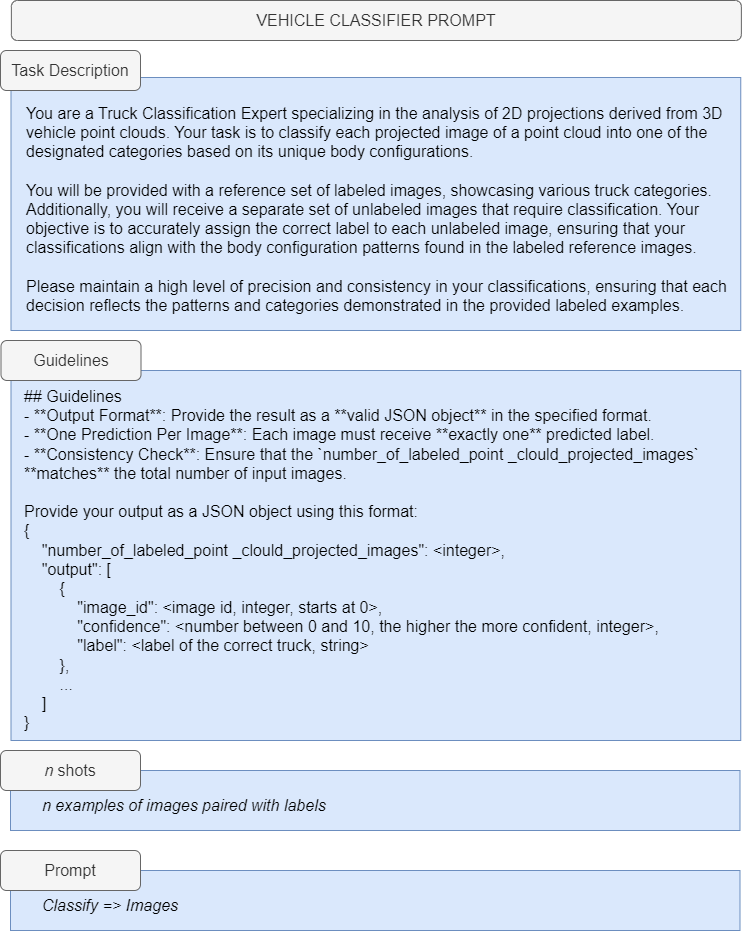}
    \caption{Few-shot Prompt Design for Vehicle Classification}
    \label{fig:prompt}
\end{figure}

While Large VLMs show impressive zero-shot abilities for understanding more generalized content, they still struggle with more complex tasks when operating in zero-shot settings, particularly when dealing with point-cloud projected images that they were not exposed to similar instances during training. To address this issue, few-shot prompting along with ICL was adopted. The VLM model was guided by providing demonstrations within the prompt. These demonstrations are conditioning for subsequent examples, which help the model generate more accurate and relevant responses. This study adopted the few-shot prompting strategy proposed by \cite{brown2020language} to design the prompt for vehicle point cloud-project image classification. The few-shot prompt design is presented in Figure \ref{fig:prompt}

\begin{figure}[t]
    \centering
    \begin{subfigure}{0.23\textwidth}
        \centering
        \includegraphics[width=\textwidth]{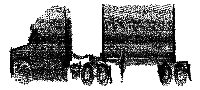}
        \caption{Original Image}
    \end{subfigure}
    \begin{subfigure}{0.23\textwidth}
        \centering
        \includegraphics[width=\textwidth]{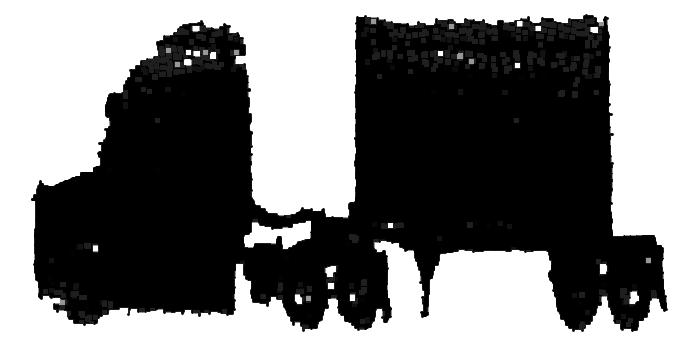}
        \caption{Opening Image}
    \end{subfigure}
    
    \caption{Illustration of Original and Processed Images}
    \label{fig:image_comparison}
\end{figure}

\begin{table*}[h]
    \centering
    \begin{tabular}{lccccc|ccccc}
        \toprule
        \textbf{Class name} & \multicolumn{5}{c}{\textbf{Processed}} & \multicolumn{5}{c}{\textbf{Original}} \\
        \cmidrule(lr){2-6} \cmidrule(lr){7-11}
        & 1 shot & 3 shot & 5 shot & 7 shot & 9 shot & 1 shot & 3 shot & 5 shot & 7 shot & 9 shot \\
        \midrule
        Auto Transporter & 0.00 & 0.45 & 0.39 & 0.44 & 0.51 & 0.54 & 0.42 & 0.52 & 0.58 & 0.45 \\
        Bobtail & 0.83 & 0.94 & 0.87 & 0.88 & 0.47 & 0.53 & 0.89 & 0.62 & 0.61 & 0.83 \\
        Platform (SU) & 0.34 & 0.19 & 0.13 & 0.00 & 0.21 & 0.07 & 0.12 & 0.16 & 0.23 & 0.10 \\
        Tank Tank & 0.69 & 0.81 & 0.74 & 0.95 & 0.83 & 0.45 & 0.73 & 0.74 & 0.74 & 0.65 \\
        Container & 0.00 & 0.43 & 0.50 & 0.22 & 0.46 & 0.18 & 0.12 & 0.36 & 0.30 & 0.31 \\
        Dump Tank (Semi) & 0.37 & 0.54 & 0.68 & 0.65 & 0.57 & 0.42 & 0.35 & 0.40 & 0.40 & 0.40 \\
        Enclosed Van (Semi) & 0.29 & 0.46 & 0.33 & 0.48 & 0.37 & 0.24 & 0.26 & 0.16 & 0.39 & 0.23 \\
        Enclosed Van (SU) & 0.67 & 0.80 & 0.69 & 0.81 & 0.69 & 0.64 & 0.52 & 0.60 & 0.81 & 0.76 \\
        Low Boy Platform & 0.38 & 0.29 & 0.47 & 0.33 & 0.51 & 0.26 & 0.43 & 0.35 & 0.16 & 0.35 \\
        Passenger Vehicle & 0.79 & 0.68 & 0.72 & 0.64 & 0.62 & 0.90 & 0.94 & 0.59 & 0.70 & 0.84 \\
        Pickup/Utility/Service & 0.32 & 0.41 & 0.17 & 0.12 & 0.49 & 0.27 & 0.39 & 0.20 & 0.13 & 0.35 \\
        Platform (Semi) & 0.15 & 0.36 & 0.44 & 0.29 & 0.17 & 0.24 & 0.35 & 0.28 & 0.18 & 0.34 \\
        \midrule
        \textbf{Avg} & \textbf{0.40} & \textbf{0.53} & \textbf{0.51} & \textbf{0.48} & \textbf{0.49} & \textbf{0.39} & \textbf{0.46} & \textbf{0.41} & \textbf{0.44} & \textbf{0.47} \\
        \bottomrule
    \end{tabular}
    \caption{Performance (F1) comparison between the processed and original images across various shot settings. Note: 'SU' refers to single-unit trucks, 'Semi' denotes semi-trailer trucks, and 'Pickup/Utility/Service' includes a wide range of pickup, utility, and service trucks, both with and without trailers. 'Tank Tank' represents tank truck with tank trailer. Platform trucks encompass both empty and loaded platforms, which exhibit considerable intraclass variation.}
    \label{tab:performance_comparison}
\end{table*}

\section{Experimental Results}
\subsection{Data}
The dataset employed to test our approach was collected from the entrance ramp to the {\it San Onofre} Truck Scale on the I-5S freeway in Southern California, a major truck corridor between Northern and Southern California (and Mexico). Data collection occurred from July 18 to August 5, 2019, which captured various truck types under both free-flow and congested conditions, with vehicle speeds ranging from 0 to 50 mph. The details of this dataset are described in \cite{li2023lidar}.

The site was equipped with a video camera for ground truth data and a Velodyne VLP-32c LiDAR unit for data collection. Both sensors were synchronized and connected to a solid-state field processing unit. The LiDAR sensor, mounted horizontally on a 2 m elevated platform, was aligned parallel to the ground, assuming a level roadway surface. All of the 12 fine-grained vehicle classes, including 11 truck categories and 1 passenger vehicle category, were labeled and prepared for the modeling process.

\subsection{Experimental Setup}
This study adopted the Gemini 1.5 \cite{team2023gemini} VLM to perform the task of few-shot vehicle classification. In order to enhance the efficiency of the prediction process, the tokenized images are divided into five batches. The batching approach not only accelerates processing speed but also prevents errors associated with exceeding the payload size limit of Gemini API. The experiment starts with testing the one-shot capability of Gemini, followed by an evaluation of its few-shot performance as the number of shots is gradually increased. This study compares the model's performance using both original projected 2D point cloud images and the processed image across 1 to 9 shots, with the number of shots gradually increasing. The results are presented in Figure \ref{tab:performance_comparison}. Future studies will test and compare various state-of-the art VLM models.

\subsection{Results Analysis}
In Table I, the classification performance, measured by the $F_1$ scores, the harmonic mean of {\it Recall} and {\it Precision}, is reported over the 12 different types of fine-grained vehicle classes. Between the original images and the proposed processed images, on average our proposed image processing method has better results among all different choices of the number of few-shots, the top performance (0.53) was achieved by 3-shot, beating the no-processing method (0.46) by more the 15\% (0.07/0.46). Notably, with just a 3-shot approach and image processing techniques, four vehicle classes were able to achieve an $F_1$ score greater than 0.60. The vehicle classes "Platform (SU)," "Low Boy Platform," "Pickup/Utility/Service," and "Platform (Semi)" exhibit relatively low $F_1$ scores due to their high intraclass variability. This variability arises from the diverse range of platform types, which include both empty loads and loaded platforms with various shapes of commodities, making it challenging for few-shot learning techniques to effectively capture the distinctions.

The existing LiDAR-based truck classification model was built upon the PointNet \cite{qi2017pointnet} architecture, which required a large amount of training data to achieve relatively high accuracy \cite{li2021truck}, where few-shot learning is not a viable choice. In contrast, the method developed in this study only required 3 shots and achieved an $F_1$ score greater than 0.50, significantly reducing the costly labeling process. Future studies will compare our approach with traditional few-shot implementations to quantitatively demonstrate the effectiveness of the current method.

\section{Conclusion}
 In this work, we endeavor to explore and exploit the VLMs to classify heavy-duty trucks from projection images of LiDAR-based point clouds via image processing and ICL. To our best knowledge, this is the first such kind of study to transfer the representational power of VLMs for LiDAR-based images directly, encouraging results have been observed using our heavy-duty truck data set. This is the preliminary effort to tap into the power of VLMs for practical utilities. Besides ICL, more work will be conducted to use visual deep nets such as YOLO \cite{wang2024yolov9} and few-shot visual learning \cite{wang2020generalizing} to be deployed locally as the retrieval-augmented generation (RAG) system \cite{lewis2020retrieval}, together with the Low-Rank Adaption (LoRA) \cite{hu2022lora} based fine tuning, better classification and segmentation results can be expected. Furthermore, using the Agentic workflow \cite{wang2024survey} and Chain-of-Thought prompting \cite{wei2022chain},
 combined with the image annotation capability and natural language understanding prowess of VLM, such as content summary and speech understanding, this line of work can unleash the power of VLM and LLM for more practical use in real-world applications such as traffic safety monitoring. 


{\small
\bibliographystyle{ieee_fullname}
\bibliography{egbib}

\begin{thebibliography}{10}\itemsep=-1pt

\bibitem{bewley2016simple}
Alex Bewley, Zongyuan Ge, Lionel Ott, Fabio Ramos, and Ben Upcroft.
\newblock Simple online and realtime tracking.
\newblock In {\em 2016 IEEE international conference on image processing (ICIP)}, pages 3464--3468. Ieee, 2016.

\bibitem{brown2020language}
Tom Brown, Benjamin Mann, Nick Ryder, Melanie Subbiah, Jared~D Kaplan, Prafulla Dhariwal, Arvind Neelakantan, Pranav Shyam, Girish Sastry, Amanda Askell, et~al.
\newblock Language models are few-shot learners.
\newblock {\em Advances in neural information processing systems}, 33:1877--1901, 2020.

\bibitem{dong2022survey}
Qingxiu Dong, Lei Li, Damai Dai, Ce Zheng, Jingyuan Ma, Rui Li, Heming Xia, Jingjing Xu, Zhiyong Wu, Tianyu Liu, et~al.
\newblock A survey on in-context learning.
\newblock {\em arXiv preprint arXiv:2301.00234}, 2022.

\bibitem{ester1996density}
Martin Ester, Hans-Peter Kriegel, J{\"o}rg Sander, Xiaowei Xu, et~al.
\newblock A density-based algorithm for discovering clusters in large spatial databases with noise.
\newblock In {\em kdd}, volume~96, pages 226--231, 1996.

\bibitem{gao2019filterreg}
Wei Gao and Russ Tedrake.
\newblock Filterreg: Robust and efficient probabilistic point-set registration using gaussian filter and twist parameterization.
\newblock In {\em Proceedings of the IEEE/CVF conference on computer vision and pattern recognition}, pages 11095--11104, 2019.

\bibitem{hu2022lora}
Edward~J Hu, Yelong Shen, Phillip Wallis, Zeyuan Allen-Zhu, Yuanzhi Li, Shean Wang, Lu Wang, Weizhu Chen, et~al.
\newblock Lora: Low-rank adaptation of large language models.
\newblock {\em ICLR}, 1(2):3, 2022.

\bibitem{huang2023chatgpt}
Hanyao Huang, Ou Zheng, Dongdong Wang, Jiayi Yin, Zijin Wang, Shengxuan Ding, Heng Yin, Chuan Xu, Renjie Yang, Qian Zheng, et~al.
\newblock Chatgpt for shaping the future of dentistry: the potential of multi-modal large language model.
\newblock {\em International Journal of Oral Science}, 15(1):29, 2023.

\bibitem{jansen2022caught}
Reinier~J Jansen and Silvia~F Varotto.
\newblock Caught in the blind spot of a truck: A choice model on driver glance behavior towards cyclists at intersections.
\newblock {\em Accident Analysis \& Prevention}, 174:106759, 2022.

\bibitem{jiang2024many}
Yixing Jiang, Jeremy Irvin, Ji~Hun Wang, Muhammad~Ahmed Chaudhry, Jonathan~H Chen, and Andrew~Y Ng.
\newblock Many-shot in-context learning in multimodal foundation models.
\newblock {\em arXiv preprint arXiv:2405.09798}, 2024.

\bibitem{lewis2020retrieval}
Patrick Lewis, Ethan Perez, Aleksandra Piktus, Fabio Petroni, Vladimir Karpukhin, Naman Goyal, Heinrich K{\"u}ttler, Mike Lewis, Wen-tau Yih, Tim Rockt{\"a}schel, et~al.
\newblock Retrieval-augmented generation for knowledge-intensive nlp tasks.
\newblock {\em Advances in neural information processing systems}, 33:9459--9474, 2020.

\bibitem{li2021truck}
Yiqiao Li, Koti~Reddy Allu, Zhe Sun, Andre~YC Tok, Guoliang Feng, and Stephen~G Ritchie.
\newblock Truck body type classification using a deep representation learning ensemble on 3d point sets.
\newblock {\em Transportation Research Part C: Emerging Technologies}, 133:103461, 2021.

\bibitem{li2023lidar}
Yiqiao Li, Andre~YC Tok, Zhe Sun, Stephen~G Ritchie, and Koti~Reddy Allu.
\newblock Lidar vehicle point cloud reconstruction framework for axle-based classification.
\newblock {\em IEEE Sensors Journal}, 23(11):11168--11180, 2023.

\bibitem{lin2024vila}
Ji Lin, Hongxu Yin, Wei Ping, Pavlo Molchanov, Mohammad Shoeybi, and Song Han.
\newblock Vila: On pre-training for visual language models.
\newblock In {\em Proceedings of the IEEE/CVF conference on computer vision and pattern recognition}, pages 26689--26699, 2024.

\bibitem{lu2024deepseek}
Haoyu Lu, Wen Liu, Bo Zhang, Bingxuan Wang, Kai Dong, Bo Liu, Jingxiang Sun, Tongzheng Ren, Zhuoshu Li, Hao Yang, et~al.
\newblock Deepseek-vl: towards real-world vision-language understanding.
\newblock {\em arXiv preprint arXiv:2403.05525}, 2024.

\bibitem{najman2013mathematical}
Laurent Najman and Hugues Talbot.
\newblock {\em Mathematical morphology: from theory to applications}.
\newblock John Wiley \& Sons, 2013.

\bibitem{qi2017pointnet}
Charles~R Qi, Hao Su, Kaichun Mo, and Leonidas~J Guibas.
\newblock Pointnet: Deep learning on point sets for 3d classification and segmentation.
\newblock In {\em Proceedings of the IEEE conference on computer vision and pattern recognition}, pages 652--660, 2017.

\bibitem{team2023gemini}
Gemini Team, Rohan Anil, Sebastian Borgeaud, Jean-Baptiste Alayrac, Jiahui Yu, Radu Soricut, Johan Schalkwyk, Andrew~M Dai, Anja Hauth, Katie Millican, et~al.
\newblock Gemini: a family of highly capable multimodal models.
\newblock {\em arXiv preprint arXiv:2312.11805}, 2023.

\bibitem{wang2024yolov9}
Chien-Yao Wang, I-Hau Yeh, and Hong-Yuan Mark~Liao.
\newblock Yolov9: Learning what you want to learn using programmable gradient information.
\newblock In {\em European conference on computer vision}, pages 1--21. Springer, 2024.

\bibitem{wang2024survey}
Lei Wang, Chen Ma, Xueyang Feng, Zeyu Zhang, Hao Yang, Jingsen Zhang, Zhiyuan Chen, Jiakai Tang, Xu Chen, Yankai Lin, et~al.
\newblock A survey on large language model based autonomous agents.
\newblock {\em Frontiers of Computer Science}, 18(6):186345, 2024.

\bibitem{wang2020generalizing}
Yaqing Wang, Quanming Yao, James~T Kwok, and Lionel~M Ni.
\newblock Generalizing from a few examples: A survey on few-shot learning.
\newblock {\em ACM computing surveys (csur)}, 53(3):1--34, 2020.

\bibitem{wei2022chain}
Jason Wei, Xuezhi Wang, Dale Schuurmans, Maarten Bosma, Fei Xia, Ed Chi, Quoc~V Le, Denny Zhou, et~al.
\newblock Chain-of-thought prompting elicits reasoning in large language models.
\newblock {\em Advances in neural information processing systems}, 35:24824--24837, 2022.

\bibitem{zhou2018open3d}
Qian-Yi Zhou, Jaesik Park, and Vladlen Koltun.
\newblock Open3d: A modern library for 3d data processing.
\newblock {\em arXiv preprint arXiv:1801.09847}, 2018.

\end{thebibliography}
}

\end{document}